\documentclass[sigconf]{aamas} 

\usepackage{balance} 

\usepackage{bm}
\usepackage{array}

\newcolumntype{R}[1]{>{\raggedleft\let\newline\\\arraybackslash\hspace{0pt}}m{#1}}

\setcopyright{ifaamas}
\acmConference[AAMAS '23]{Proc.\@ of the 22nd International Conference
on Autonomous Agents and Multiagent Systems (AAMAS 2023)}{May 29 -- June 2, 2023}
{London, United Kingdom}{A.~Ricci, W.~Yeoh, N.~Agmon, B.~An (eds.)}
\copyrightyear{2023}
\acmYear{2023}
\acmDOI{}
\acmPrice{}
\acmISBN{}

\acmSubmissionID{502}

\title[MIRL-ToM]{Multiagent Inverse Reinforcement Learning via\\ Theory of Mind Reasoning}

\author{Haochen Wu}
\affiliation{
  \institution{University of Michigan\\ Dept. of Mechanical Engineering}
  \city{Ann Arbor, MI}
  \country{United States}}
\email{haochenw@umich.edu}

\author{Pedro Sequeira}
\affiliation{
  \institution{SRI International \\ Artificial Intelligence Center}
  \city{Menlo Park, CA}
  \country{United States}}
\email{pedro.sequeria@sri.com}

\author{David V. Pynadath}
\affiliation{
  \institution{University of Southern California Institute for Creative Technologies}
  \city{Los Angeles, CA}
  \country{United States}}
\email{pynadath@usc.edu}

\begin{abstract}
We approach the problem of understanding how people interact with each other in collaborative settings, especially when individuals know little about their teammates, via Multiagent Inverse Reinforcement Learning (MIRL), where the goal is to infer the reward functions guiding the behavior of each individual given trajectories of a team’s behavior during some task. Unlike current MIRL approaches, we \emph{do not} assume that team members know each other's goals a priori; rather, that they collaborate by adapting to the goals of others perceived by observing their behavior, all while jointly performing a task. To address this problem, we propose a novel approach to MIRL via Theory of Mind (MIRL-ToM). For each agent, we first use ToM reasoning to estimate a posterior distribution over baseline reward profiles given their demonstrated behavior. We then perform MIRL via decentralized equilibrium by employing single-agent Maximum Entropy IRL to infer a reward function for each agent, where we simulate the behavior of other teammates according to the time-varying distribution over profiles. We evaluate our approach in a simulated 2-player search-and-rescue operation where the goal of the agents, playing different roles, is to search for and evacuate victims in the environment. Our results show that the choice of baseline profiles is paramount to the recovery of the ground-truth rewards, and that MIRL-ToM is able to recover the rewards used by agents interacting both with known and unknown teammates.
\end{abstract}

\newpage

\newcommand{\BibTeX}{\rm B\kern-.05em{\sc i\kern-.025em b}\kern-.08em\TeX}

\usepackage{algorithm, algorithmic}
\usepackage{amsmath}
\usepackage{multirow}
\usepackage{colortbl}
\usepackage{caption}
\usepackage{subcaption}
\usepackage{enumitem}

\begin{document}


\pagestyle{fancy}
\fancyhead{}


\maketitle


\section{Introduction}
\begin{figure*}[!t]
    \centering
    \includegraphics[width=0.78\textwidth]{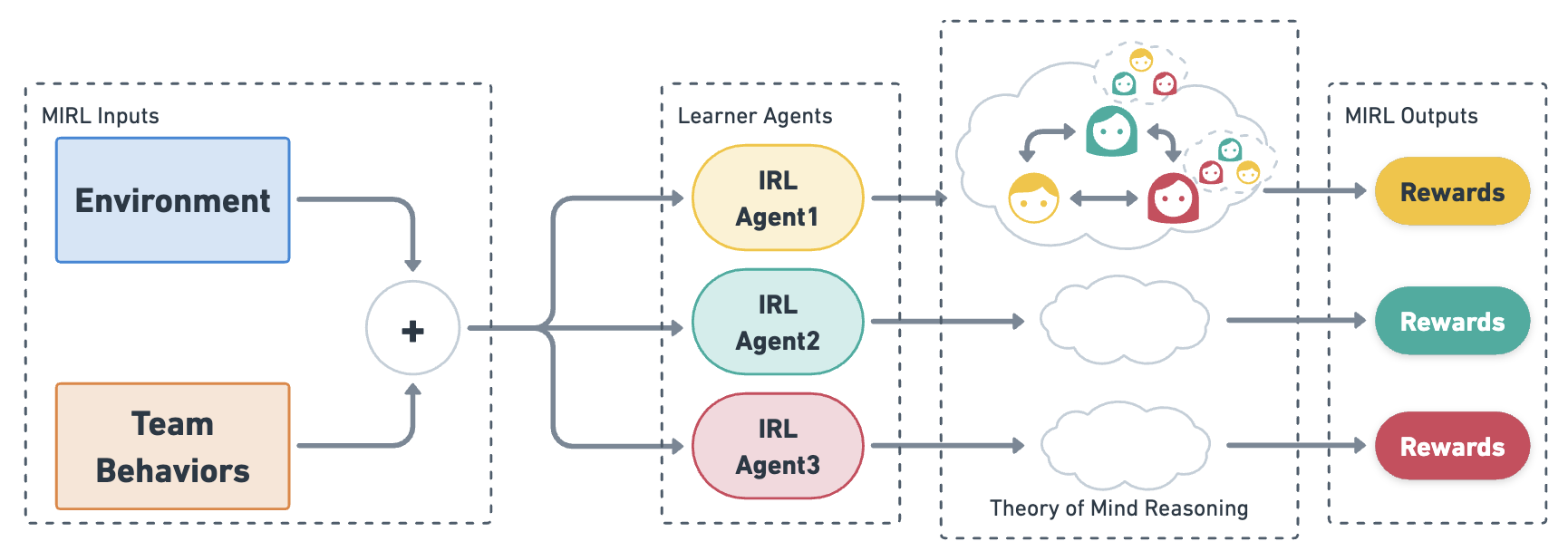}
    \vspace{-3mm}
    \caption{Overview of the proposed approach for Multiagent Inverse Reinforcement Learning via Theory of Mind reasoning (MIRL-ToM). Given the environment and the demonstrated behaviors of a team of humans, each learner agent performs single-agent IRL by reasoning about other agents' behavior via ToM and output the resulting reward functions.}
    \label{fig:MIRLwToM}
\end{figure*}

Understanding the underlying strategies of how humans interact with each other in collaborative teaming scenarios is a vital component of creating systems that appropriately predict team behavior, understand collaboration dynamics, and provide assistance to potentially improve performance.
This is particularly challenging when people are collaborating for the first time and therefore know little about the goals, strategies, and intentions of others, yet have to adapt to each other as they perform a cooperative task. 

One way to predict what individuals will do is to create models of their decision-making process given traces of their behavior, using machine learning techniques. One approach is to learn a policy, i.e., a mapping from states to actions, directly from observation. This includes imitation learning techniques, e.g., behavior cloning \cite{ALVINN} that has the power of replicating the behavior of others without the need for environment and dynamics modeling \cite{BehaviorClone}. However, such approaches lack the capability of understanding the underlying intentions of individuals or the goals in the task. Further, in teamwork settings, knowing the policy of individuals alone makes it hard to understand how they coordinate their behavior \cite{CoordStrategy}, or whether there are impediments to cooperation, e.g., a mismatch between the preferences of each team member.

An alternative, more robust method is to assume that the behavior of each individual is the result of a Reinforcement Learning (RL) process, where agents learn how to perform a task through trial-and-error interactions with a dynamic and  uncertain environment \cite{RLIntro}. The goal of an RL agent is specified by a reward function that encodes its preferences and whose output the agent wants to maximize over its lifetime. Therefore, understanding the underlying strategy of an individual by observing its behavior relies on recovering the reward function guiding the behavior during task performance. This problem is known as Inverse Reinforcement Learning (IRL) \cite{IRLALgo, survey} and involves inferring the reward function used by an individual, referred to as the \emph{expert}, given traces of its behavior. IRL has advantages compared to imitation learning approaches in that the reward function is the most succinct and robust representation of the expert's goals and allows for the prediction of behavior in novel situations or different tasks (generalization) \cite{IRLALgo}.

Within IRL, apprenticeship learning \cite{ApprenticeLearning} established a linear programming method to match the learned reward model to the feature counts derived from state-action pairs in the expert's demonstrations. A game-theoretic approach \cite{GameTheoryAppLearn} was also proposed to learn from playing repeated games even in the absence of experts. Maximum margin IRL \cite{MaxMargin} formulates a quadratic problem to find the reward solution producing behavior paths that are closer to the expert's behavior than others by a margin that scales with the path difference loss. One of the main challenges of IRL that is addressed by these approaches is that the IRL problem is ill-posed --- the same observed behavior can be explained by multiple reward functions, and the same reward function can lead to different behaviors under stochastic dynamics \cite{survey}. To resolve this ambiguity, many IRL algorithms provide probabilistic formulations by finding the reward function that maximizes the likelihood of the demonstrated trajectories. In particular, Maximum entropy (MaxEnt) IRL  \cite{MaxEntIRL} finds a trajectory distribution where the expert demonstrations are exponentially preferred. Bayesian IRL \cite{BIRL} presents an efficient sampling method to infer the posterior distributions of reward function parameters.

Understanding the strategies and intentions underlying the behavior of a team in collaborative tasks requires extending IRL to multiagent settings, which is known as Multiagent Inverse Reinforcement Learning (MIRL). In MIRL, the goal is to infer the individual reward functions guiding the behavior of each agent, given trajectories of a team's (joint) behavior. It is often assumed that the joint behavior is the result of an equilibrium solution concept, such as Nash equilibrium \cite{NashEqui, MAIRLStrategy}. This makes the MIRL problem harder because, in addition to multiple reasonable reward functions given the demonstrations, there may be multiple equilibria that are a solution to a combination of individual rewards \cite{MAIRLGames, MAIRLStrategy}. 

Many of the existing approaches to MIRL assume the existence of a centralized learner that computes the joint action equilibrium given the (known) reward functions of each individual \cite{MAIRLDecNonCoop, MAIRLGames, MAIRLComp}. This implies that all individuals perfectly know each other's preferences (task payoffs/reward functions) at all times and then play according to the equilibrium solution. However, assuming perfect information about others in a team might be too restrictive. First, the individuals might be interacting with unknown teammates, e.g., the observed behavior was produced by individuals recruited to perform a collaborative task who tried to adapt to each other's behaviors while performing the task. Second, the assumptions formed by each individual regarding the teammates' goals and intentions, created by observing their behavior, might be incorrect due to the noisy nature of human behavior. Finally, because individuals are adapting their behavior to that of their team members (non-stationarity), the assumptions about their preferences must constantly be updated as a result of the individuals' interaction. All of this makes it less likely that a team's demonstrated behavior was derived from a unique, stationary, known a priori equilibrium.

To address the aforementioned issues, we borrow the concept Theory of Mind (ToM) from cognitive science \cite{ToM} which ascribes mental states to explain and predict the actions of others. In particular, we model each individual as adapting to the behavior of the teammates by maintaining a probability distribution over \emph{baseline behavior profiles} for each other member of the team, which is updated as the task unfolds and interactions between team members occur. Computationally, these profiles correspond to domain-specific reward models encoding typified preferences in a task. We then split MIRL into two phases (see overview in Fig.~\ref{fig:MIRLwToM}): 
\begin{enumerate}
    \item \emph{Model inference:} for each individual in the team, we use ToM reasoning to estimate a posterior distribution over baseline profiles for each other teammate given their behavior included in the trajectories --- this models how each team member might have modeled each other by observing their behavior during task performance.
    \item \emph{Decentralized MIRL:} We perform MIRL via a decentralized equilibrium approach where we break down the MIRL problem into multiple single-agent IRL problems similar to the approach in \cite{MAIRLStrategy}. A reward function is inferred given the corresponding individual's demonstrated behavior via MaxEnt IRL, where we simulate the behavior of other team members according to the time-varying distribution over baseline profiles computed in Phase 1 --- this models best-response strategies conditioned on the perceived models of others.
\end{enumerate}

The analogy between IRL and ToM has been drawn in \cite{ToMIRL} to link the mental states of humans to the reward models of IRL agents \cite{CogToM}. Despite applications of ToM in multiagent RL \cite{MachineToM}, limited research has been done to investigate the underlying strategic behaviors and motivations of teams in the MIRL setting. The main contribution of this paper is a decentralized MIRL framework using ToM reasoning (MIRL-ToM) that extends MaxEnt IRL to multiagent settings. Our framework allows modeling interactions between arbitrarily unknown teammates under the assumption of a set of baseline reward profiles. We deal with uncertainties in the intentions of other teammates by performing ToM reasoning over the profiles in a Bayesian setting. 

We evaluate our approach in a simulated 2-player search-and-rescue task where each agent plays a different role with a distinct set of skills in the task. We designed rewards for each agent that include both task- and social-related aspects allowing for different types of coordination strategies to emerge. We assess the capability of our MIRL-ToM approach in recovering the original reward functions of agents given different sets of baseline profiles. Overall, our results show that the choice of baseline profiles in Phase 1 is paramount to the recovery of ground-truth rewards in phase 2. In particular, including baseline profiles corresponding to the opposite goals of team-members prevents the convergence of MaxEnt IRL, while including baselines whose goals are close to those of the original teammates leads to reward functions similar to the ground-truth. Further, we show that our method, using ToM reasoning over the demonstrated behaviors, can be used to recover the rewards used by agents interacting both with known and unknown teammates, which attests to the robustness of our approach.


\section{Related Work}
Generalizing IRL to multiagent systems is hard, because we aim to learn individual reward functions that explain both how each agent behaves---which involves recovering the rewards via IRL---and how together the agents coordinate---that involves modeling the joint behavior as a solution to an equilibrium concept. Within MIRL, in \cite{MAIRL} an algorithm was developed to train a centralized agent that optimizes coordination of behaviors for traffic-routing problems; however, while learning the rewards via IRL, the agents ignore their teammates which are treated as part of the environment. In \cite{MAIRLGames}, Bayesian inference is performed on zero-sum two-person games by optimizing rewards under a likelihood function that encodes the notion of a minimax equilibrium, but the solution is not scalable to $n$-player games and complex domains. 
Further, many MIRL solutions rely on computing some notion of equilibrium given the individual reward functions at each iteration. Some approaches deal with competitive games by learning Markov perfect equilibrium policies given individual rewards \cite{MAIRLDecNonCoop}, playing against Nash Equilibrium policies through adversarial training \cite{MAIRLComp}, evaluating undesirable strategic behavior by extracting utility functions from demonstrations \cite{MAIRLStrategy}, and leveraging Generative Adversarial Networks \cite{GAN} to retrieve reward functions in high-dimensional domains \cite{MA-AIRL}. However, these methods still require finding an equilibrium policy throughout learning, and it is assumed that experts have perfect knowledge of the goals and intentions of their teammates, ignoring the dynamic nature of interactions and the possibility of first encounters. The proposed MIRL-ToM, combining ToM reasoning with single-agent MaxEnt IRL, is able to not only decentralize the model reasoning and reward learning process but also considers situations of unknown teammates and adaptation via belief updates over a set of baseline reward profiles.


\section{Background}
\subsection{Reinforcement Learning}
Reinforcement learning (RL) \cite{RLIntro} relies on the formalism of (single-agent) Markov Decision Process (MDP), which is defined as a tuple of $\left<S, A, T, R, \gamma\right>$, where $s\in S$ is a finite set of states, $a\in A$ is a finite set of actions, $T:=Pr(s',a,s)$ is the transition probability between states given an action, $R:=R(s,a)$ is the reward of reaching a state and performing an action, and $\gamma\in[0,1)$ is the discount factor that encourages receiving larger reward at earlier steps. Under the Bellman equations, the optimal quality value of an action given a state satisfies $Q^*(s,a)=R(s,a)+\gamma\mathbb{E}_{s'\sim T}[maxQ^*(s',a')]$. (Forward) RL algorithms find the optimal policy $\pi:S\rightarrow A$ that can be retrieved using $\pi^*=argmax_a Q^*(s,a)$. 

\subsection{Inverse Reinforcement Learning}
Inverse Reinforcement Learning (IRL) \cite{IRLALgo} represents the inverse problem of RL, i.e., finding the reward function $R$ of the learner agent given $S, A, T$ and a collection of demonstrated trajectories $\mathcal{D}=\{\zeta_0\dots\zeta_i\dots\}$ following the internal policy of an expert. Each trajectory $\zeta_i=\{(s_0, a_0)\dots (s_j, a_j) \dots (s_t, a_t)\}$ consists of state-action pairs over $t$ steps. IRL aims to find the reward function parameters that maximize the likelihood of demonstrations. Maximum Entropy IRL \cite{MaxEntIRL} (MaxEnt IRL) utilizes the principle of maximum entropy and assumes that the demonstrated action given a state is exponentially preferred over other actions. MaxEnt IRL considers the trajectory reward $R(\zeta)$ as a linear combination of state feature counts (i.e., $R(\zeta)=\theta^T\phi(\zeta)$) and iteratively finds the parameters $\theta^*=\arg\max_\theta {\mathbb{E}_{\zeta\sim Pr(\zeta|\theta)}[\sum_j r(s_j, a_j|\theta)]}$ that best reproduce the demonstrated behaviors such that the probability of taking a trajectory $\zeta$ is proportional to the exponential of the rewards along the path (i.e., $Pr(\zeta|\theta)\propto e^{R(\zeta)}$).

\subsection{Multiagent Inverse Reinforcement Learning}
Multiagent Inverse Reinforcement Learning (MIRL) extends IRL to multiagent settings. The collected team trajectories are denoted as $\mathcal{D}=\{\zeta_0\dots\zeta_i\dots\}$, where $\zeta_i=\{(s_0, \hat{a}_0)\dots (s_j, \hat{a}_j) \dots (s_t, \hat{a}_t)\}$ is each team trajectory in the form of state-joint action pairs $(s_j, \hat{a}_j)$ over $t$ steps, where $\hat{a}^k_j$ denotes the action of agent $k$ at step $j$. MIRL then aims to recover a set of reward function parameters for all agents $\Theta^*=\{\theta^*\}=\arg\max_\Theta{\mathbb{E}_{\zeta\sim Pr(\zeta|\Theta)}[\sum_j r(s_j, \hat{a}_j|\Theta)]}$ that best match the collected team trajectories.

\subsection{PsychSim}
We implemented our approach using PsychSim \cite{PsychSim}, a framework that models decision-theoretic agents with recursive Theory-of-Mind (ToM) for interactive social simulation \cite{PsychSim}. To achieve this, PsychSim models world dynamics under the Multiagent Partially-Observable Markov Decision Process (MPOMDP) formalism \cite{MPOMDP, MPOMDP2}. At each step, each agent selects an action based on the values computed for each available action by assessing risk/reward trade-offs in decisions up to a specified planning horizon (cumulative discounted reward), conditioned on the actions of others. Namely, during planning each agent computes the action values for all other agents using recursive ToM, where others are considered reward maximizers under the (mental) models that the agent has of them. To model uncertainty, agents can form beliefs about the state of the environment and the models of other agents, which are updated via Bayesian inference based on observations of state features and the actions of others, respectively.


\section{Multiagent Inverse Reinforcement Learning via Theory of Mind}
\begin{figure}[!t]
    \centering
    \includegraphics[width=0.35\textwidth]{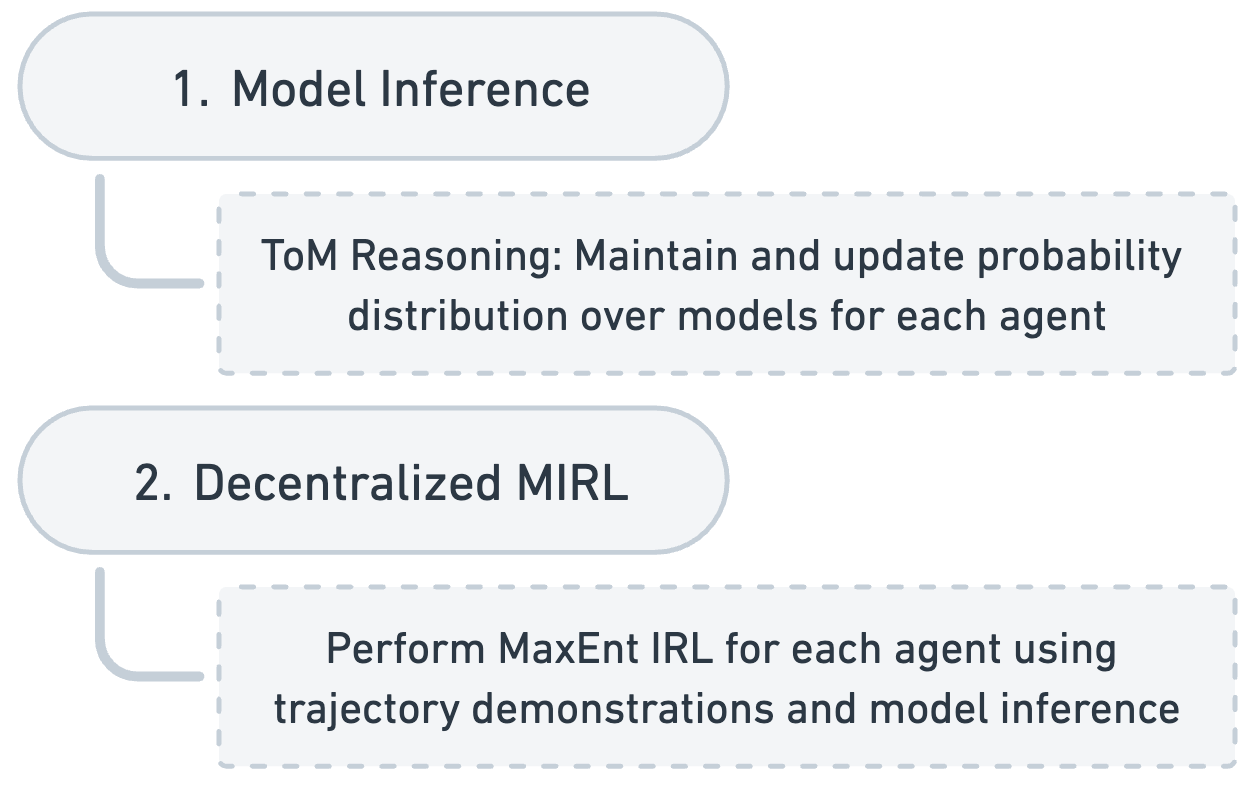}
    \vspace{-2mm}
    \caption{Two phases of the Multiagent Inverse Reinforcement Learning via Theory of Mind} 
    \label{fig:MIRL_Process}
\end{figure}

We assume that individuals in a team coordinate their behavior without explicit communication. In such cases, they need to reason about the goals and intentions of others while computing their own actions. We also assume the existence of a set of \emph{baseline agent profiles} that encode domain-specific, notional preferences in the task of interest. For example, in a search-and-rescue task, this might correspond to having profiles for trying to search for victims, triage them, calling for backup, etc. Before a task, an individual might assume that teammates behave according to a prior distribution over the baseline profiles, e.g., uniform, or a preference given to a profile given background information. As the task progresses and individuals observe the behavior of others, this distribution is updated accordingly. We propose Multiagent Inverse Reinforcement Learning via Theory of Mind (MIRL-ToM) to allow reasoning about others' behavior while performing a task, corresponding to a decentralized approach to learn the reward functions of each team member. The approach consists of two phases as shown in Fig. \ref{fig:MIRL_Process}, the details of which are discussed in the following sections.

\subsection{Model Inference}

In the model inference phase, we use ToM reasoning to update a probability distribution (belief) over baseline profiles given the behavior of an agent in the trajectories. Before the inference process, we pre-select the set of baseline profiles for each agent. In general, the profiles can encode any agent decision-making \emph{model}, and we denote the initial profile set by $\{m_i\}$. In this paper, we consider profiles to correspond to predefined reward functions, i.e., reward profiles. Let us denote by $b^k(m)$ the prior distribution that an agent holds over the models of another agent $k$. For each step of each trajectory $\zeta_i$, $b^k(m)$ is updated via Bayesian inference:
\begin{equation}\label{eq:update_model}
    b'(m)\propto Pr(a|m,\zeta)\times b(m|\zeta)
\end{equation}
where we drop the agent indexing $k$ for convenience. This phase augments the team trajectories, $\hat{\mathcal{D}}=\{\hat{\zeta}_i\}$, by adding the agent's belief over agent models at each step, resulting in a set of state-joint action-model distribution tuple sequences: $\hat{\zeta}_i=\{(s_j, \hat{a}_j, b_j(m)\}$. In loose terms, this phase captures how the intentions of agents could be modeled by others given their actions in the trajectories, conditioned on a set of agent profiles and the prior distribution.%
\footnote{We note that the initial set of profiles and the prior distribution need not to be the same for all observing agents. Similarly, the belief update process can be different for distinct agents observing the same teammate, e.g., by restricting observability to local interactions between agents.}

\subsection{Decentralized MIRL}

After each learner agent computes the (time-varying) beliefs about the expected reward models of its teammates, we perform decentralized MIRL by computing a reward function for each individual agent separately. Without loss of generality, here we assume that the reward functions $R(s,a):=\theta^T\phi(s,\hat{a})$ are linear combinations of (potentially nonlinear) reward features $\phi(s,\hat{a})$ parameterized by reward weights $\theta$. We follow the maximum entropy principle to find the individual reward function where the demonstrated trajectories of the agent are exponentially preferred \cite{MaxEntIRL}, i.e.:
\begin{equation}
    \label{eq:obj}
    \theta^*
    = \arg\max_\theta{L\left(\theta|b(m)\right)}
    = \arg\max_\theta{\sum_{\zeta_i\in\mathcal{D}} \log Pr\left(\zeta_i|\theta,b(m)\right)}
\end{equation}

\begin{algorithm}[b]
    \caption{Decentralized MIRL via ToM}
    \begin{algorithmic}[1]
        \label{alg:DecMAMaxEntIRL}
        \STATE $\mathcal{D} \leftarrow$ Collection of team trajectories
        \FOR{each learner agent $k$}
        \STATE $b_0(m)$: prior model distribution, $\phi(s,a)$: feature vector
        \STATE $\hat{\mathcal{D}}=\{\hat{\zeta}_i\} \leftarrow$ Model inference on $\mathcal{D}$ using ToM reasoning
        
        \STATE $\phi^k_{emp} \leftarrow \sum_{\zeta_i\in\mathcal{D}}\sum_{(s_j,\hat{a}^k_j)\in\zeta_i}\phi(s_j,\hat{a}^k_j)$ 
        \STATE Initialize reward vector weights $\theta^k$
        \WHILE{$\theta^k$ not converging}
        \STATE $\pi^k(a|s,b(m),\theta^k) \leftarrow$ forward pass 
        \STATE $\phi^k_{est} \leftarrow \sum_{\zeta_i\sim\pi}\sum_{(s_j,\hat{a}^k_j|b_j(m))\in\zeta_i} Pr(s_j|\pi^k,\theta^k)\phi(s_j,\hat{a}^k_j)$
        \STATE $\nabla L(\theta^k) \leftarrow \phi^k_{est} - \phi^k_{emp}$
        \STATE $\theta^k \leftarrow \theta^k - \alpha\nabla L(\theta^k)$
        \ENDWHILE
        \ENDFOR
        \RETURN $\{\theta^k\}$
    \end{algorithmic}
\end{algorithm}

\begin{figure}[!t]
    \centering
    \includegraphics[width=0.4\textwidth]{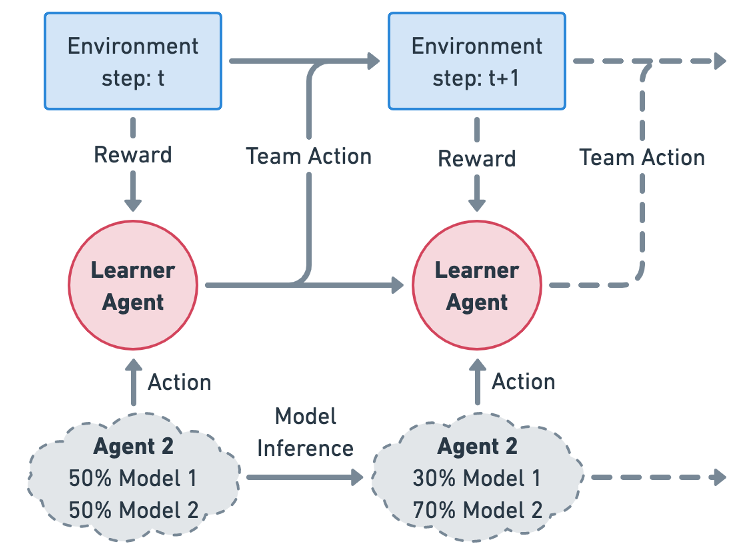}
    \caption{Forward pass of feature count estimation. Each learner agent uses the model distribution data to select the actions of other agents and select an action given the current reward model. The feature counts are computed from the environment states and the team actions.}
    \label{fig:FC_Est}
\end{figure}

We propose an algorithm that extends the MaxEnt IRL \cite{MaxEntIRL} to the MIRL setting via decentralized equilibrium computation, where at each iteration, we approximate the distribution over paths of an agent conditioned on both the current reward weight and the belief over agent models computed in the model inference phase, as shown in Alg.~\ref{alg:DecMAMaxEntIRL}. For ease of explanation, in Alg.~\ref{alg:DecMAMaxEntIRL} we show the case where the team contains only two agents, i.e., there is only one model distribution, $b(m)$, to be computed.
To solve the MIRL problem, we learn a reward weight vector $\theta^k$ for each individual agent $k$ by iteratively improving according to whether the weight vector leads to the behavior similar to that exhibited by the corresponding expert in the demonstrated trajectories, as measured by comparing the estimated feature counts (FCs) against the empirical FCs (Alg.~\ref{alg:DecMAMaxEntIRL}, line 5). Therefore, we achieve a decentralized equilibrium by having each individual reward function be the solution to the best response strategy to the perceived models of the other agents during estimated FCs computation. We note that, as an advantage to this approach, the learning process for each agent can be completely distributed and parallelized.
At each iteration, we assume the existence of a forward RL solver computing the policy $\pi^k$ given the current reward weight vector candidate, $\theta^k$, where the actions of the other agent are optimized according to the distribution over models $b(m)$ at each timestep (Alg.~\ref{alg:DecMAMaxEntIRL}, line 8). During the gradient descent step (Alg.~\ref{alg:DecMAMaxEntIRL}, lines 9-11), the estimated FCs $\phi_{est}$ are computed by sampling Monte Carlo trajectories using the resulting joint policy. This process can be visualized in Fig.~\ref{fig:FC_Est}.

\section{Results}
To assess the effectiveness of the proposed MIRL-ToM approach, we consider a collaborative environment and a team of two agents. Instead of evaluating our approach with real human data, where the underlying reward functions cannot be accessed, here we specify a set of ground-truth rewards, collect trajectories by having the agents maximize those rewards according to some predefined model of others, and then compare the learned to the ground-truth reward functions for each agent. This is a necessary first step to assess the capabilities (and limitations) of our approach before testing with human data.%
\footnote{Our implementation and experiments are available at \href{https://github.com/usc-psychsim/mirl-tom-aamas23}{https://github.com/usc-psychsim/mirl-tom-aamas23}.}

\subsection{Experiment}

\subsubsection{Environment}
\label{sec:env}

\begin{figure}[!t]
    \centering
    \includegraphics[width=0.4\textwidth]{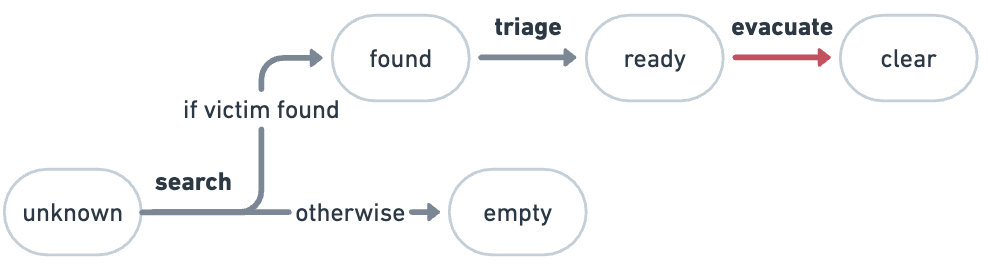}
    \caption{Transition dynamics of the victim status. $\bm{search}$ either finds a victim or an empty cell, $\bm{triage}$ prepares a victim and requires two agents at the same location to $\bm{evacuate}$ it.}
    \label{fig:state_transition}
\end{figure}

We designed a collaborative multiagent search-and-rescue (S\&R) operation using PsychSim \cite{PsychSim} which allows the modeling of partial observations, stochastic state-action dynamics, and ToM reasoning via agent modeling.
In our S\&R task, two agents need to cooperate in order to search all locations in a gridworld environment and evacuate all victims as quickly as possible. Initially, the victims' locations are unknown to the agents, which we model via a uniform belief over the existence of victims in each location, which is set to the real value once agents enter the corresponding cell. The state of the environment $s\in S:=S_1^P\times\dots\times S_n^P$ describes a gridworld with $n$ locations, where each location has a feature $v\in V:=\{unknown, found, ready, clear, empty\}$ denoting the victim status at that location. The action set includes three types of actions: 1) movement actions: $\{up, down, left, right, wait\}$ that deterministically move the agent in the corresponding direction, 2) victim-handling actions: $\{search, triage, evacuate\}$ that changes the victim status feature, and 3) communication action, $call$, which allows an agent to incentivize others to come to its location. The state transition of the victims' status is deterministic, the dynamics of which are illustrated in Fig.~\ref{fig:state_transition}. As denoted by the red arrow, to model the need for cooperation, we require the two agents to be at the same location and perform action $evacuate$ at the same time for a victim to be evacuated.

\begin{table}[ht]
    \caption{Ground-truth reward functions for the two agents as linear combinations of features $\bm{\phi_i}$ weighted by $\bm{\theta_i}$.}
    \begin{tabular}{l l l l}
    \textcolor{orange!40}{$\blacksquare$} & \hspace{-3mm}Role-Related Tasks & \textcolor{cyan!20}{$\blacksquare$} & \hspace{-3mm}Social/Helping
    \end{tabular}
    \begin{subtable}{0.48\textwidth}
        \centering
        \begin{tabular}{|p{0.15\textwidth}|p{0.08\textwidth}|p{0.61\textwidth}|}
             \hline
             \rowcolor[gray]{0.8}
             \multicolumn{1}{|c|}{$\bm{\phi_i(s,a)}$} & \multicolumn{1}{c|}{$\bm{\theta_i}$} & \textbf{Description} \\
             \hline
            \rowcolor{orange!30}
             Dist2Vic & 0.06 & Distance to the nearest $found$ victim\\
             \rowcolor{cyan!10}
             Search & 0.06 & Whether to search $unknown$ locations\\
             \rowcolor{orange!30}
             Triage & 0.19 & Whether to triage $found$ victims\\
             \rowcolor{orange!30}
             Evacuate & 0.63 & Whether to evacuate $ready$ victims\\
             \rowcolor{cyan!10}
             Wait & 0.03 & Whether to standby\\
             \rowcolor{cyan!10}
             Call & 0.03 & Whether to call for help if needed\\
             \hline
        \end{tabular}
    \label{tb:medic_rwdfeat}
    \caption{Medic}
    \end{subtable}

    \begin{subtable}{0.48\textwidth}
    \centering
        \begin{tabular}{|p{0.15\textwidth}|p{0.08\textwidth}|p{0.61\textwidth}|}
             \hline
             \rowcolor[gray]{0.8}
             \multicolumn{1}{|c|}{$\bm{\phi_i(s,a)}$} & \multicolumn{1}{c|}{$\bm{\theta_i}$} & \textbf{Description} \\
             \hline
            \rowcolor{cyan!10}
             Dist2Help & \hspace{1mm}0.25 & Distance to the agent with $call$ action\\
             \rowcolor{orange!30}
             Search & \hspace{1mm}0.25 & Whether to search $unknown$ locations\\
             \rowcolor{cyan!10}
             Evacuate & \hspace{1mm}0.50 & Whether to evacuate $ready$ victims\\
             \hline
        \end{tabular}
    \label{tb:explorer_rwdfeat}
    \caption{Explorer}
    \end{subtable}
    \label{tb:rwdfeat}
    \vspace{-10mm}
\end{table}

\subsubsection{Ground-Truth Agents (Experts)}
\label{sec:gt_agents}

We consider a team of two agents with designated roles---Medic and Explorer---each defining different action sets and reward functions. The Medic agent has the full action set, while the Explorer agent is not able to $wait$, $triage$, or $call$, i.e., it can only \emph{explore} the environment. The reward function of each agent $R(s,a):=\sum_i\theta_i \phi_i(s,a)$ is linearly parameterized by ground truth weights $\theta$ as defined in Table~\ref{tb:rwdfeat}. The distance features compute the Manhattan distance from the agent to the target location. The action-related features are indicator functions that denote whether an action was taken in the current state. As seen in Table~\ref{tb:rwdfeat}, reward features are categorized into role-related \emph{task} features (orange) and \emph{social} features (blue). We model the Medic as prioritizing the rescue of victims, while the Explorer prioritizes searching but also reacts to the call for help from the Medic.

\subsection{Collecting and Visualizing Trajectories}
\label{sec:collect_trajs}

\begin{figure}[!t]
    \centering
    \includegraphics[width=0.47\textwidth]{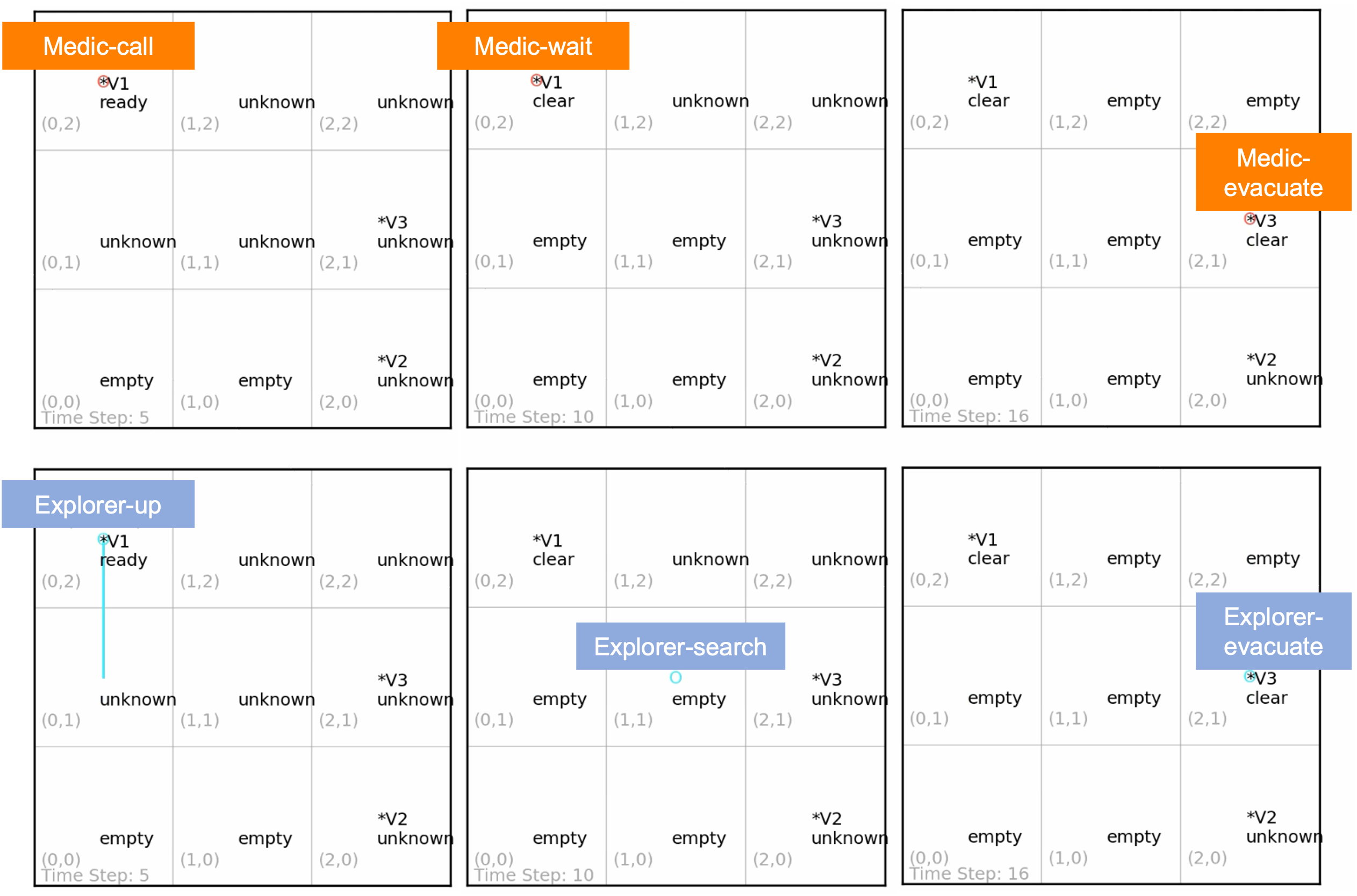}
    \caption{An example trajectory at timestep $\bm{t=4}$ (left), $\bm{t=10}$ (center), and $\bm{t=16}$ (right) for the Medic (top) and Explorer (bottom) agents.}
    \label{fig:episode}
\end{figure}

Given the ground-truth reward functions defined in Table~\ref{tb:rwdfeat}, we start with the case where the two agents interact with each other and perform the task knowing exactly each other's reward function (perfect models). Notwithstanding, we note that the MIRL-ToM procedure is unaware of this and is only given the generated team trajectories and a set of baseline profiles. The goal is to test the importance of different priors over profiles for the recovery of the original rewards. 
We collected $16$ trajectories, each with $25$ steps, for the team of agents, produced by having the agents plan with a horizon of $2$ using a soft-max action selection. Fig. \ref{fig:episode} shows the collaborative behavior of two agents at three different steps in one example trajectory. At $t=4$, the Medic calls for help, and the Explorer moves toward the Medic instead of searching or moving away; $t=10$ illustrates role-dependent actions: the Medic is waiting, because all victims are either unknown or clear, while the Explorer keeps searching other locations; at $t=16$, both agents evacuate the victim together.%
\footnote{Example simulated animations showing the agents interacting in the S\&R environment are \href{https://github.com/usc-psychsim/mirl-tom-aamas23/blob/main/traj_animation.zip}{included HERE} as the supplementary material.}

\subsection{Model Inference Results}
\label{sec:model_inference}

Based on the reward features $\phi_i$ detailed in Sec.~\ref{sec:gt_agents}, we designed a set of baseline profiles as shown in Table~\ref{tb:rwdmodels} by creating different variations of the ground-truth rewards and the task- and social-oriented features. The goal was to create a set of reward functions one would typically expect individuals to follow in our S\&R task.

From these reward profiles, we define the following experimental conditions, i.e., each defining the set of agent models/profiles used in the Model Inference phase (see Sec.~\ref{sec:model_inference}):

\begin{figure}[!t]
    \centering
    \includegraphics[width=0.48\textwidth]{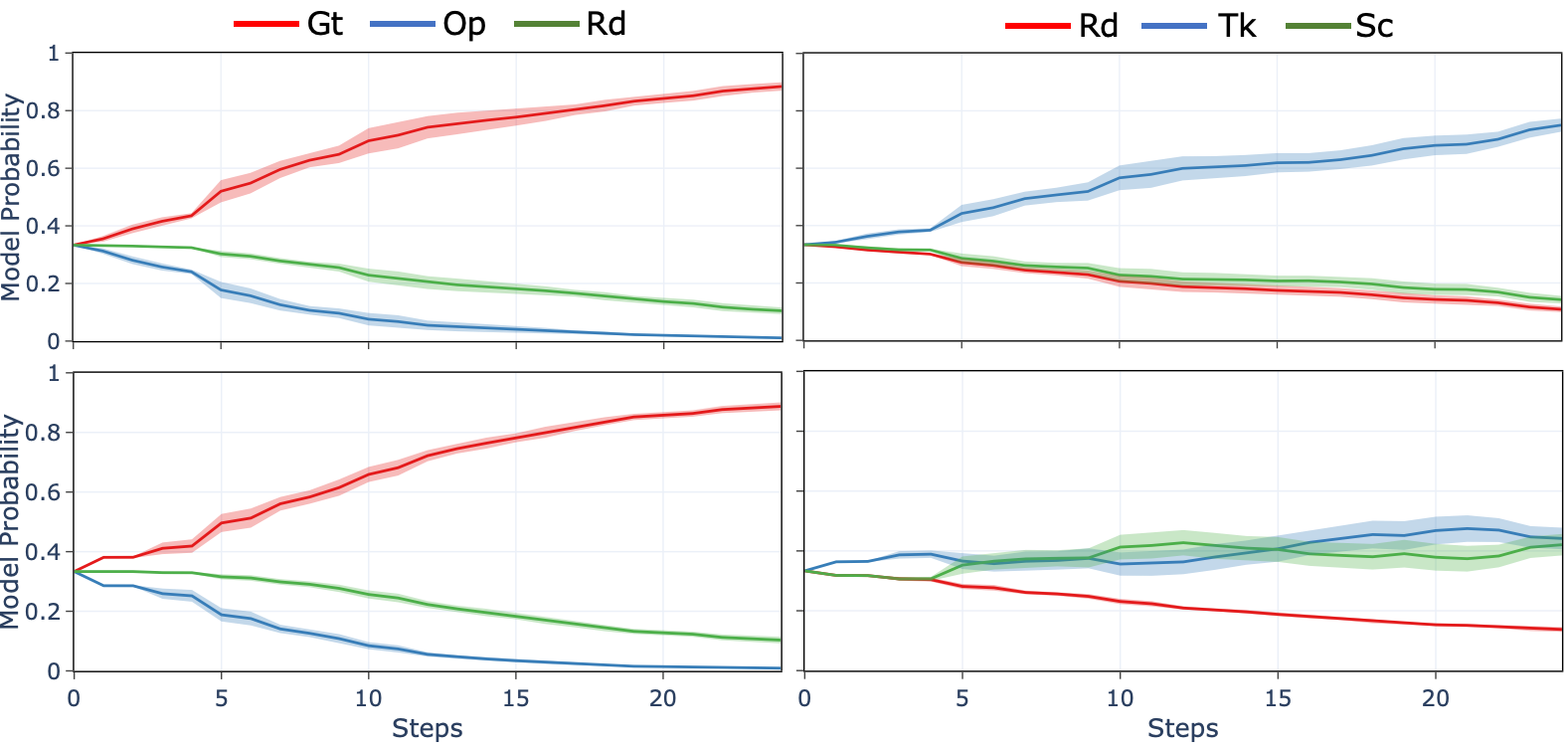}
    \vspace{-3mm}
    \caption{Average model probabilities with standard deviations (shaded) of 16 trajectories for Medic (top) and Explorer (bottom) agents under two experimental conditions: $\bm{cond}$-3 (Gt-Op-Rd, left) and $\bm{cond}$-4 (Rd-Tk-Sc, right).} 
    \label{fig:model_inference_cond34}
\end{figure}
\begin{table}[h]
    \centering
    \caption{Baseline reward profiles}
    \vspace{-3mm}
    \begin{tabular}{|c|c|p{0.25\textwidth}|}
         \hline
         \rowcolor[gray]{0.8}
          & \textbf{Reward Function} & \textbf{Description} \\
         \hline
         Gt & $\theta_{gt}$ & Ground truth rewards\\
         Op & $-\theta_{gt}$ & Opposite goals \\
         Rd & $\theta_0$ & No goals / random behavior \\
         Tk & $[\theta_{task},\theta_{social}=0]$ & Focus on task-specific aspects\\
         Sc & $[\theta_{task}=0,\theta_{social}]$ & Focus on social aspects\\
         \hline
    \end{tabular}
\label{tb:rwdmodels}
\vspace{-5mm}
\end{table}
\begin{description}
    \item[$\bm{cond}$-1: Gt,] always believe the other agent uses Gt rewards.
    \item[$\bm{cond}$-2: Op,] always believe the other agent uses Op rewards.
    \item[$\bm{cond}$-3: Gt-Op-Rd,] update belief $b(\{Gt,Op,Rd\})$ via ToM.
    \item[$\bm{cond}$-4: Rd-Tk-Sc,] update belief $b(\{Rd,Tk,Sc\})$ via ToM.
\end{description}

Condition $cond$-1 corresponds to the setting where the learner agent uses the ground-truth rewards of the other (perfect model) to compute the estimated FCs during MaxEnt IRL. The rationale is to assess the best-response (decentralized equilibrium) computation during single-agent IRL. $cond$-2 presents a situation where the learner agent always believes its teammate has goals opposite to those of the ground-truth agent. Here we assess the importance of agent co-dependencies in the task and test the limits of MIRL in recovering the original rewards given contradictory models of others. For these first two conditions, the model distribution is fixed so we do not perform model inference via ToM reasoning. In contrast, $cond$-3 and $cond$-4 both use ToM reasoning. $cond$-3 verifies the accuracy of the model inference procedure in identifying the ground truth model given the observed behavior, while $cond$-4 investigates whether MIRL-ToM can still recover rewards close to the original without being given the Gt model given, i.e., such as it might happen when analyzing real human data.

\begin{figure*}[!t]
    \centering
    \includegraphics[width=0.65\textwidth]{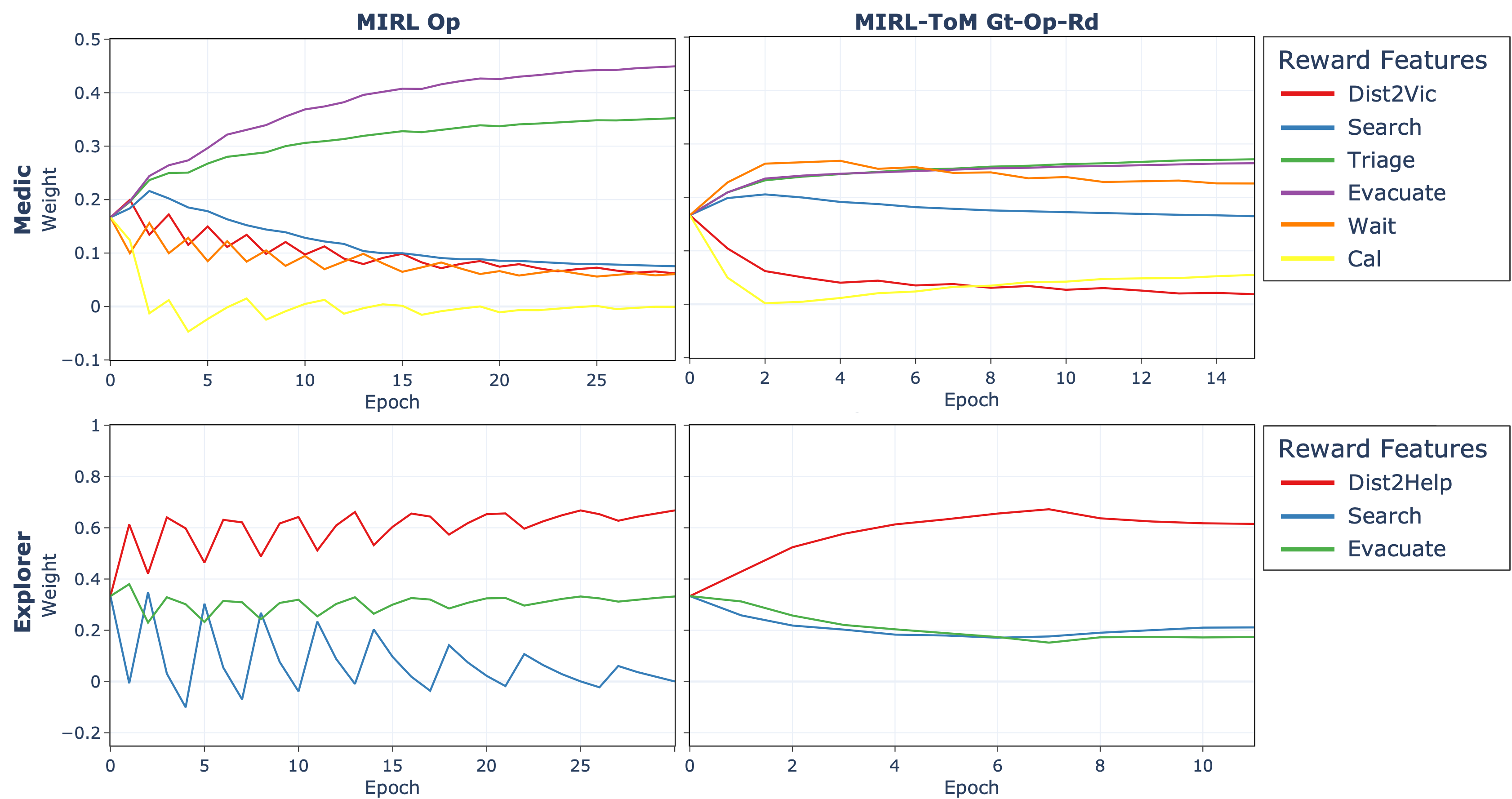}
    \caption{Learning curves of the reward weights during IRL for Medic (top) and Explorer (bottom) in two experimental conditions: $\bm{cond}$-2 (left) and $\bm{cond}$-3 (right).} 
    \label{fig:reward_converge}
\end{figure*}

For $cond$-3 and $cond$-4, we perform Bayesian inference using PsychSim by computing the posterior distribution over agent models given their behavior at each step of the $16$ collected trajectories using Eq.~\ref{eq:update_model}. The mean model probabilities at each timestep are shown in Fig.~\ref{fig:model_inference_cond34}. In $cond$-3 (Fig. \ref{fig:model_inference_cond34} left), the Gt models can clearly be identified apart from the Rd models, meaning that the observed behavior does denote different agent strategies (otherwise, a random behavior could be equally probable). In addition, the Op model has a very low associated probability at the end of trajectories. For $cond$-4 (Fig. \ref{fig:model_inference_cond34} right), the Medic agent is identified as being more task-oriented, which makes sense since it has available the full action set. However, for the Explorer agent, it is harder to differentiate between the Tk and Sc models, meaning that the ground-truth behavior is perceived as being equally task- and social-oriented.

\subsection{Decentralized MIRL Results}
\label{sec:recover}

We applied the proposed decentralized MIRL-ToM approach in each of the aforementioned experimental conditions, with initially uniform reward weights ($\theta_0$), maximum $30$ IRL training epochs, and a learning rate ($\alpha$) exponential decay of $0.9$. For the forward pass (Alg.~\ref{alg:DecMAMaxEntIRL}, line 8), we collected $16$ Monte Carlo trajectories in PsychSim to estimate the FCs. 

\begin{table}[h]
    \vspace{-1mm}
    \centering
    \caption{Learned reward functions for the two agents, for all conditions in the known teammates setting.}
    \vspace{-3mm}
    \begin{subtable}{0.48\textwidth}
        \centering
        \begin{tabular}{|p{0.14\textwidth}|rr|rc|rr|}
             \hline
             \rowcolor[gray]{0.6}
             \multicolumn{1}{|c|}{\textbf{Feature}} & \multicolumn{2}{c|}{\textbf{Profile} $\bm{\theta_i}$} & \multicolumn{2}{c|}{\textbf{MIRL} $\bm{\theta_i}$} & \multicolumn{2}{c|}{\textbf{MIRL-ToM} $\bm{\theta_i}$}\\
             \hline
             \rowcolor[gray]{0.9}
             \multicolumn{1}{|c|}{$\phi_i(s,a)$} & \multicolumn{1}{c}{Gt} & \multicolumn{1}{c|}{Op} & \multicolumn{1}{c}{$cond$-1} & \multicolumn{1}{c|}{$cond$-2} & \multicolumn{1}{c}{$cond$-3} & \multicolumn{1}{c|}{$cond$-4} \\
             \hline
             Dist2Vic    & 0.06 & -0.06 & 0.02 & - & 0.02 & 0.00\\
             Search      & 0.06 & -0.06 & 0.18 & - & 0.16 & 0.16\\
             Triage      & 0.19 & -0.19 & 0.23 & - & 0.27 & 0.29\\
             Evacuate    & 0.63 & -0.63 & 0.25 & - & 0.26 & 0.28\\
             Wait        & 0.03 & -0.03 & 0.27 & - & 0.23 & 0.26\\
             Call        & 0.03 & -0.03 & 0.06 & - & 0.06 & 0.01\\
             \hline
        \end{tabular}
        \label{tb:medic_rwdweight}
        \caption{Medic}
    \end{subtable}
    \begin{subtable}{0.48\textwidth}
        \centering
        \begin{tabular}{|p{0.14\textwidth}|rr|rc|rr|}
             \hline
             \rowcolor[gray]{0.6}
             \multicolumn{1}{|c|}{\textbf{Feature}} & \multicolumn{2}{c|}{\textbf{Profile} $\bm{\theta_i}$} & \multicolumn{2}{c|}{\textbf{MIRL} $\bm{\theta_i}$} & \multicolumn{2}{c|}{\textbf{MIRL-ToM} $\bm{\theta_i}$}\\
             \hline
             \rowcolor[gray]{0.9}
             \multicolumn{1}{|c|}{$\phi_i(s,a)$} & \multicolumn{1}{c}{Gt} & \multicolumn{1}{c|}{Op} & \multicolumn{1}{c}{$cond$-1} & \multicolumn{1}{c|}{$cond$-2} & \multicolumn{1}{c}{$cond$-3} & \multicolumn{1}{c|}{$cond$-4} \\
             \hline
             Dist2Help & 0.25 & -0.25 & 0.01 & - & 0.62 & 0.26\\
             Search    & 0.25 & -0.25 & 0.23 & - & 0.21 & 0.43\\
             Evacuate  & 0.50 & -0.50 & 0.76 & - & 0.17 & 0.30\\
             \hline
        \end{tabular}
        \label{tb:explorer_rwdweight}
        \caption{Explorer}
        \vspace{-10mm}
    \end{subtable}
    \label{tb:rwdweight}
\end{table}

The learned reward functions inferred via MIRL-ToM for each condition are summarized in Table~\ref{tb:rwdweight}. We observe that the reward weights obtained for $cond$-3 and 4 using ToM reasoning are very close to those of $cond$-1, where the Gt model is used for both agents. Especially for $cond$-4, we see that model inference via ToM captured the behavior of the teammates accurately enough to allow for the recovery of the original behavior. Some results in this table can better be understood by looking at the learning curves, i.e., the evolution of the reward weights $\theta_i$ at each IRL iteration. In particular, Fig.~\ref{fig:reward_converge} shows the curves for $cond$-2 and $cond$-3. As can be seen, the reward weights converge to their final values approximately within $15$ training epochs except for $cond$-2, where agents always believe the other agent uses the Op model. The reward weights for this condition were not able to converge (Fig. \ref{fig:reward_converge}, left), either oscillating or not converging within the maximum number of iterations.%
\footnote{The seemingly converging curves are due to the decreased learning rate $\alpha$.}
Because the agent uses an incorrect model of its teammate, the resulting coordination behavior, computed via the best-response strategy learned by the agent given the current rewards and the other's model distribution (Alg.~\ref{alg:DecMAMaxEntIRL}, line 8) leads to feature counts that are inconsistent with the empirical ones (lines 9-10), which in turn leads to a dramatic change in the reward vectors (line 11) in the next iteration.

\begin{table}[t]
    \vspace{-1mm}
    \caption{Similarity of recovered behavior for the two agents, for all conditions in the known teammates setting.}
    \vspace{-3mm}
    \begin{subtable}{0.48\textwidth}
        \centering
        \begin{tabular}{|p{0.14\textwidth}|rr|r|rr|}
             \hline
             \rowcolor[gray]{0.6}
             \multicolumn{1}{|c|}{\textbf{Feature}} & \multicolumn{2}{c|}{\textbf{Profile}} & \multicolumn{1}{c|}{\textbf{MIRL}} & \multicolumn{2}{c|}{\textbf{MIRL-ToM}}\\
             \hline
             \rowcolor[gray]{0.9}
             \multicolumn{1}{|c|}{$\phi_i(s,a)$} & \multicolumn{1}{c}{Gt} & \multicolumn{1}{c|}{Op} & \multicolumn{1}{c|}{$cond$-1} & \multicolumn{1}{c}{$cond$-3} & \multicolumn{1}{c|}{$cond$-4} \\
             \hline
             \multicolumn{6}{|c|}{\textbf{Empirical Feature Counts}}\\
             \hline
             Dist2Vic    & 11.86 & 0.00 & 11.53 & 15.39 & 14.48\\
             Search      &  2.40 & 0.00 &  1.75 &  3.13 &  2.06\\
             Triage      &  2.97 & 0.00 &  2.91 &  2.97 &  2.94\\
             Evacuate    &  2.94 & 0.00 &  2.81 &  2.63 &  2.66\\
             Wait        &  8.97 & 0.00 & 10.88 &  6.28 &  7.78\\
             Call        &  4.78 & 0.00 &  2.72 &  4.38 &  4.44\\
             \hline
             \multicolumn{6}{|c|}{\textbf{Similarity Metrics}}\\
             \hline
              FC Diff   & 2.38 & 30.55 & 4.52 & \textbf{7.53} & \textbf{4.50}\\
             \hline
              $\pi$ Div & 0.00 &  0.25 & 0.13 & \textbf{0.10} & \textbf{0.13}\\
             \hline
        \end{tabular}
        \label{tb:medic_sim_metrics}
        \caption{Medic}
    \end{subtable}
    \begin{subtable}{0.48\textwidth}
        \centering
        \begin{tabular}{|p{0.14\textwidth}|rr|r|rr|}
             \hline
             \rowcolor[gray]{0.6}
             \multicolumn{1}{|c|}{\textbf{Feature}} & \multicolumn{2}{c|}{\textbf{Profile}} & \multicolumn{1}{c|}{\textbf{MIRL}} & \multicolumn{2}{c|}{\textbf{MIRL-ToM}}\\
             \hline
             \rowcolor[gray]{0.9}
             \multicolumn{1}{|c|}{$\phi_i(s,a)$} & \multicolumn{1}{c}{Gt} & \multicolumn{1}{c|}{Op} & \multicolumn{1}{c|}{$cond$-1} & \multicolumn{1}{c}{$cond$-3} & \multicolumn{1}{c|}{$cond$-4} \\
             \hline
             \multicolumn{6}{|c|}{\textbf{Empirical Feature Counts}}\\
             \hline
             Dist2Help & 6.50 & 0.00 & 2.41 & 6.45 & 4.24\\
             Search    & 6.59 & 0.00 & 7.06 & 6.34 & 7.38\\
             Evacuate  & 2.94 & 0.00 & 2.81 & 2.63 & 2.66\\
             \hline
             \multicolumn{6}{|c|}{\textbf{Similarity Metrics}}\\
             \hline
             FC Diff & 2.25 & 14.84 & 3.06 & \textbf{2.13} & \textbf{1.26}\\
             \hline
             $\pi$ Div & 0.00 & 0.32 & 0.02 & \textbf{0.03} & \textbf{0.04}\\
             \hline
        \end{tabular}
        \label{tb:explorer_sim_metrics}
        \caption{Explorer}
    \vspace{-10mm}
    \end{subtable}
    \label{tb:sim_metrics}
\end{table}

As shown in Table~\ref{tb:rwdweight}, the reward weights recovered through IRL do not exactly match the ground-truth rewards (Gt column). Still, they allow us to recover the relative importance that some features have on the agents' behavior, e.g., dealing with victims (including waiting for the Explorer for evacuation) is more important for the Medic agent, while searching for and evacuating victims are the priorities of the Explorer agent. As discussed earlier, different reward functions can lead to similar behavior, so a better way to investigate the similarities between the learned and expert behavior is by using similarity metrics. Here we use the following:
\begin{description}
    \item[FC Diff:] feature count difference between a policy using the IRL rewards vs. demonstrated behavior: $\sum_i |\phi_i^\theta - \phi_i^\mathcal{D}|$.
    \item[$\bm{\pi}$ Div:] mean Jensen-Shannon Divergence\cite{fuglede2004jsd} between the action distributions of the ground-truth and learned policy, over all steps in all trajectories: $\overline{JSD}(\pi(a_j|s_j;\theta), \pi(a_j|s_j;\theta_{gt}))$.
\end{description}
The results are summarized in Table~\ref{tb:sim_metrics}. The Profile columns correspond to the best (Gt) and worst (Op) behavior similarities achieved. The FC Diff for Gt denotes the expected error in sampling policies using the same rewards but different random seeds (hence not zero). As the number of simulated trajectories (currently 16) increases, the FC Diff would approach to zero. The MIRL column shows very similar feature counts and behavior, which is again expected because it corresponds to using a fixed Gt model of the teammate (no ToM). As for the MIRL-ToM columns, they show that our method can achieve very similar behavior (highlighted in bold), which denotes the importance of the model inference phase. Interestingly, more similar behavior was attained in $cond$-4, which does not include the Gt rewards in the model distribution, although we did not assess the statistical significance of this result. We also note that $\pi$ Div is not as close to zero due to the observed difference in the reward weights in Table~\ref{tb:rwdweight}, and the probabilities are dependent on the reward weight magnitude.

\subsection{Unknown Teammates Setting}
\label{sec:uncertain_teammates}

\begin{table}[!t]
    \caption{Similarity of recovered behavior for the two agents in the unknown teammates setting.}
    \vspace{-3mm}
    \begin{subtable}{0.48\textwidth}
    \centering
        \begin{tabular}{|p{0.14\textwidth}|R{0.2\textwidth}|R{0.2\textwidth}|}
             \hline
             \rowcolor[gray]{0.6}
             \multicolumn{1}{|c|}{\textbf{Feature}} & \multicolumn{1}{c|}{\textbf{Demo}}& \multicolumn{1}{c|}{\textbf{MIRL-ToM}}\\
             \hline
             \rowcolor[gray]{0.9}
             \multicolumn{1}{|c|}{$\phi_i(s,a)$} & \multicolumn{1}{c|}{$\theta_{gt}$} & \multicolumn{1}{c|}{$\theta_{irl}$} \\
             \hline
             \multicolumn{3}{|c|}{\textbf{Empirical Feature Counts}}\\
             \hline
             Dist2Vic    & 14.05 & 14.44\\
             Search      &  1.99 &  2.16\\
             Triage      &  2.49 &  2.97\\
             Evacuate    &  2.68 &  2.81\\
             Wait        &  7.13 &  7.88\\
             Call        &  5.88 &  5.25\\
             \hline
             \multicolumn{3}{|c|}{\textbf{Similarity Metric}}\\
             \hline
             FC Diff & - & 2.55 \\
             \hline
        \end{tabular}
        \label{tb:medic_sim_metrics_unknown}
        \caption{Medic}
    \end{subtable}
    \begin{subtable}{0.48\textwidth}
    \centering
        \begin{tabular}{|p{0.14\textwidth}|R{0.2\textwidth}|R{0.2\textwidth}|}
             \hline
             \rowcolor[gray]{0.6}
             \multicolumn{1}{|c|}{\textbf{Feature}} & \multicolumn{1}{c|}{\textbf{Demo}}& \multicolumn{1}{c|}{\textbf{MIRL-ToM}}\\
             \hline
             \rowcolor[gray]{0.9}
             \multicolumn{1}{|c|}{$\phi_i(s,a)$} & \multicolumn{1}{c|}{$\theta_{gt}$} & \multicolumn{1}{c|}{$\theta_{irl}$} \\
             \hline
             \multicolumn{3}{|c|}{\textbf{Empirical Feature Counts}}\\
             \hline
             Dist2Help & 6.49 & 7.23 \\
             Search    & 8.25 & 6.50 \\
             Evacuate  & 2.68 & 2.81 \\ 
             \hline
             \multicolumn{3}{|c|}{\textbf{Similarity Metric}}\\
             \hline
             FC Diff & - & 2.63 \\
             \hline
        \end{tabular}
        \label{tb:explorer_sim_metrics_unknown}
        \caption{Explorer}
    \end{subtable}
    \label{tb:sim_metrics_unknown}
    \vspace{-10mm}
\end{table}

In this section, we consider demonstrations that are collected from a team of agents who form beliefs about each other's goals and update them based on observations of their behavior \emph{during} task performance. We again use PsychSim to generate the demonstrated trajectories, but now the expert agents infer the models of each other given a uniform prior over three reward profiles: Gt, Op and Rd.
We applied MIRL-ToM by performing model inference given the demonstrations followed by MaxEnt IRL for each agent. In Table~\ref{tb:sim_metrics_unknown}, we compare the empirical feature counts, computed from the demonstrations (column Demo) where agents use the ground-truth rewards, $\theta_{gt}$, with the feature counts produced by the agents using the reward weights resulting from MIRL-ToM, $\theta_{irl}$.  

The results show that even in the case where the demonstrations are generated by agents that have imperfect models of their teammates, our approach can recover similar behavior. By performing model inference, we are able to model the process whereby the individuals ``discover'' about and adapt to their teammates' behavior.


\section{Conclusions}
We proposed an approach to Multiagent Inverse Reinforcement Learning using Theory of Mind reasoning (MIRL-ToM). Unlike other approaches to MIRL, our method does not assume full knowledge of teammates' reward functions while performing the task to compute the equilibrium strategy. Rather, we use ToM to reason about how each individual might have modeled the intentions of others by observing their behavior throughout the task, given a set of baseline reward profiles. We then break down MIRL into single-agent MaxEnt IRL for each individual, where we infer the reward functions guiding their behavior conditioned on the inferred distribution over profiles. Ascribing mental states and reward models of teammates not only accounts for imperfect knowledge about individual behavior and uncertain strategies of others but also enables a computationally efficient approach to compute the equilibrium strategy in a decentralized manner. The proposed MIRL-ToM shows the ability to recover similar behavior in terms of trajectory feature counts in both known- and unknown-teammate cases. In addition, although we experimented with a 2-player collaborative scenario, our framework can be applied to $n$-player, general-sum problems.

Our preliminary experiments were conducted using synthetic expert demonstration data to show the validity and effectiveness of MIRL-ToM. A natural next step is to apply the method to data collected from humans to provide insights into which factors---task-, social-, and emotional-related---affect their behavior the most in collaborative settings. Further, we will explore more interpretable reward structures that do not rely on linear relationships between basis features. We are also interested in exploring ways to create the baseline profiles, because the effectiveness of MIRL-ToM seems to rely heavily on the initial set. One option is to create an iterative approach, whereby the learned reward functions in one iteration are added to the set of baseline profiles until convergence.



\begin{acks}
This material is based upon work supported by the Defense Advanced Research Projects Agency (DARPA) under Contract No. W911NF-20-1-0011. Any opinions, findings and conclusions or recommendations expressed in this material are those of the author(s) and do not necessarily reflect the views of the Defense Advanced Research Projects Agency (DARPA).
\end{acks}



\bibliographystyle{ACM-Reference-Format} 
\bibliography{references}

\end{document}